\documentclass[english,letterpaper,10pt,conference]{ieeeconf}
\usepackage[T1]{fontenc}
\usepackage[latin9]{inputenc}
\usepackage{float}
\usepackage{amsmath}
\usepackage{graphicx}
\usepackage[numbers]{natbib}

\makeatletter

\providecommand{\tabularnewline}{\\}

\IEEEoverridecommandlockouts
\usepackage{url}
\usepackage{indentfirst}

\usepackage[footnotesize]{caption}

\usepackage{balance}
\usepackage{algorithm,algpseudocode}
\usepackage{color}

\@ifundefined{showcaptionsetup}{}{%
 \PassOptionsToPackage{caption=false}{subfig}}
\usepackage{subfig}
\makeatother

\usepackage{babel}
\begin{document}
\title{\textbf{Acoustic Localization and Communication Using a MEMS
Microphone for Low-cost and Low-power Bio-inspired Underwater Robots}}
\author{Akshay Hinduja$^{1}$, Yunsik Ohm$^{1}$, Jiahe Liao$^{2}$, Carmel Majidi$^{1}$, and Michael Kaess$^{2}$\thanks{The authors are with the Department of Mechanical Engineering\textsuperscript{1}, the Robotics Institute\textsuperscript{2}, Carnegie Mellon University, Pittsburgh, PA 15213, USA. {\tt\small \{ahinduja, yohm, jiahel, cmajidi, kaess\}@andrew.cmu.edu }}
\thanks{This work was partially supported by the National Oceanographic Partnership Program under ONR award N00014-18-1-2843.}
\thanks{https://github.com/rpl-cmu/underwater-acoustic-pseudoranging\textsuperscript{3}}}

\maketitle
\begin{abstract}
Having accurate localization capabilities is one of the fundamental requirements of autonomous robots. For underwater vehicles, the choices for effective localization are limited due to limitations of GPS use in water and poor environmental visibility that makes camera-based methods ineffective. 
Popular inertial navigation methods for underwater localization using Doppler-velocity log sensors, sonar, high-end inertial navigation systems, or acoustic positioning systems require bulky expensive hardware which are incompatible with low-cost, bio-inspired underwater robots. In this paper, we introduce an approach for underwater robot localization inspired by GPS methods known as acoustic pseudoranging. Our method allows us to potentially localize multiple bio-inspired robots equipped with commonly available micro electro-mechanical systems microphones. This is achieved through estimating the time difference of arrival of acoustic signals sent simultaneously through four speakers with a known constellation geometry. We also leverage the same acoustic framework to perform one-way communication with the robot to execute some primitive motions. To our knowledge, this is the first application of the approach for the on-board localization of small bio-inspired robots in water. Hardware schematics and the accompanying code are released to aid further development in the field\textsuperscript{3}.
\end{abstract}

\section{Introduction and Related Work\label{sec:Introduction-and-Related}}

Underwater robots are increasingly being used to explore uncharted depths, inspect artificial undersea structures, as well as monitor aquatic life and the physical and chemical properties of their surrounding environment. While larger robots like autonomous underwater vehicles (AUVs) are more suited to tasks such as the inspection of ship hulls, harbors~\citep{Teixeira16iros} and sub-sea pipelines, bio-inspired robots are better equipped to closely monitor underwater life given their smaller size and lack of moving parts likely to disturb the surrounding environment~\cite{raj2016fish,paley2021bioinspired}. 
Bio-inspired robots have improved over recent years in aspects of locomotion, control, and communication. In 2014, Marchese et al.~\citep{Marchese2014AutonomousSR} presented their novel actuation method for a soft robotic fish capable of agile movement. Soon after in 2015, the group presented a compact acoustic communication module for in-water control of the robot~\citep{DelPreto2015wuwnet}. More recently, an improvement over the previous works was presented by Katzschmann et al.~\citep{Katzschmann18science} through their soft robotic fish, SoFi, which could be controlled using a handheld acoustic modem based controller presented in the research done by Marchese et al.~\citep{DelPreto2015wuwnet}. The drawback here is that the controller is considerably expensive and can only communicate with one robot at a time. A more recent example of a smaller bio-inspired underwater robot is PATRICK from Patterson et al.~\citep{Patterson20iros}, which demonstrated the capability of performing closed-loop locomotion planning using an external camera and Bluetooth communication. Here, the use of Bluetooth for passing instructions to the robot limits it to only operate at surface level or shallow water since it requires a non-submerged Bluetooth antenna to receive instructions. However, despite the limitations, these examples demonstrate the promising transition towards autonomy for bio-inspired underwater robots.

\begin{figure}[H]
\begin{centering}
\subfloat[Setup for pseudorange localization]{\centering{}\includegraphics[width=0.95\columnwidth]{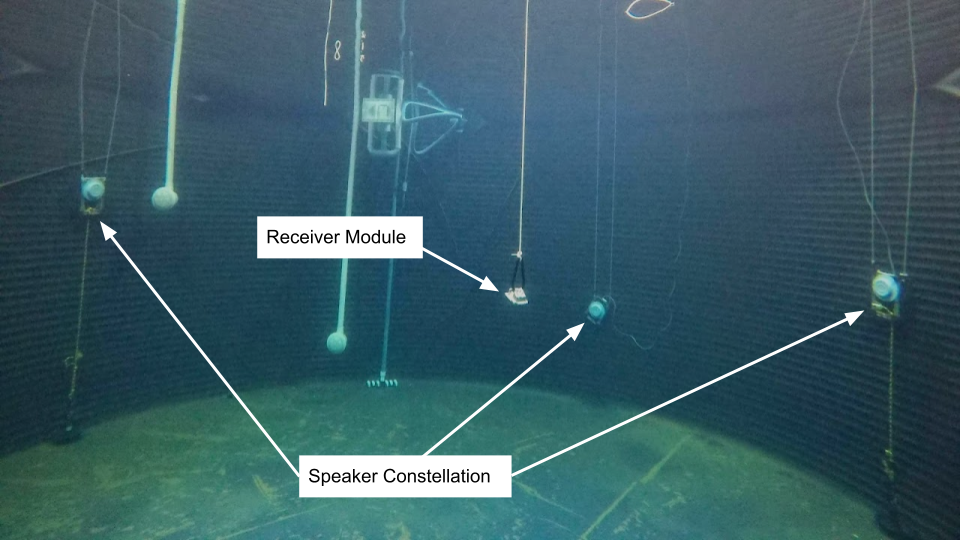}}\medskip{}
\subfloat[Bio-inspired fish robot]{\centering{}\includegraphics[width=0.95\columnwidth]{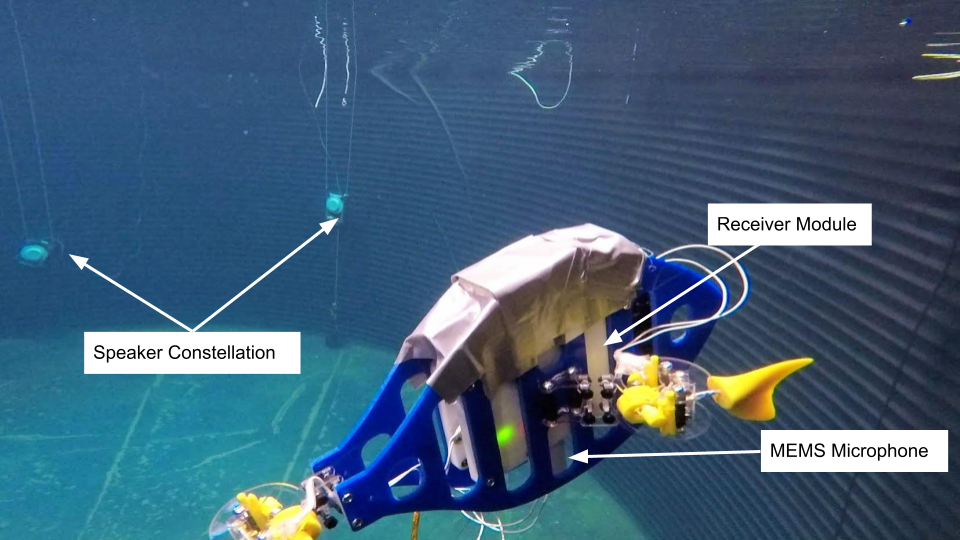}}
\par\end{centering}
\caption{Water Tank Experiments: (a) Experimental verification of acoustic pseudoranging based localization. Estimated positions are compared to known ground truth positions. (b) A proof-of-concept bio-inspired robot with three actuated fins connected to the receiver module is verified to execute selective motions on receiving specific acoustic signals. }\label{fig:experiment_pics}

\end{figure}

A fundamental necessity for an autonomous robot to navigate and perform specific tasks is the ability to localize itself accurately. Traditionally in robotics, there are systems that can achieve accurate location estimates by utilizing one or more sensors like cameras, lidars, radars, inertial measurement units (IMUs) and the global positioning system (GPS). Underwater robots face unique challenges when it comes to localization. In real-world scenarios, sensors like cameras and lidar are ineffective due to turbidity, and signals from absolute positioning systems like GPS cannot be received underwater. Most commercial AUVs rely on sensors like Doppler velocity logs (DVLs), depth sensors and IMUs for inertial navigation methods to help localize themselves. Sonars and cameras can be used for aiding inertial navigation by leveraging the geographical features of the surrounding environment. As mentioned earlier, cameras are effective only in clear water and sonars often make the cost of underwater vehicles prohibitively expensive. Costs and practicality aside, their weights and bulky form factors limit them to be used only on larger robots. In this work, our focus is on localization for  small, low-cost and low-power, untethered robots. 

For localization of smaller and cheaper underwater robots, such as bio-inspired robot fish, acoustic positioning methods have more potential as a viable solution. The oldest and most popular methods of acoustic positioning are long baseline or short baseline (LBL/SBL) and ultra short baseline (USBL) systems~\citep{Paull2014AUVNav,Kinsey2006IFAC}. These methods are two-way travel-time (TWTT) methods, which means the beacons and the AUVs need to be equipped with active acoustic systems to communicate with one another. The use of atomic clocks to synchronize both beacons and AUVs can allow for localization using one-way travel-time (OWTT) methods. Earlier approaches achieved range-only OWTT localization and navigation by using filtering or fusion with other onboard sensors~\citep{Eustice2007icra,Jakuba2015oceans}. More recently, Rypkema et al.~\citep{Rypkema2017icra} presented their novel method of one-way travel-time inverted ultra-short baseline (OWTT-iUSBL), which allowed for real-time on-board navigation using a single speaker and an array of four passively listening hydrophones on the AUV. Matched filtering and phased-array beamforming on the four received signals from each hydrophone gives a range and bearing estimate of the robot with respect to the speaker position. 

While the OWTT-iUSBL can perform effective localization with a relatively cost effective setup compared to commercial AUVs, the components used are still expensive and too heavy to be used on much smaller bio-inspired robots. Atomic clocks, which make OWTT methods possible, are still expensive relative to other options. For a swarm of robots, these costs can add up significantly. Accurate time-of-arrival information is paramount for OWTT methods, and cheaper embedded real time clocks (RTCs) are prone to significant drift and lack the superior precision of atomic clocks. Time-difference of arrival (TDOA) techniques, on the other hand, allow us to avoid the need to know the exact time a signal was sent. This is achieved by instead measuring the differences between the arrival time of multiple signals which were known to be transmitted simultaneously from different beacons. This technique is commonly referred to as \textit{pseudoranging}, and is the foundation of GPS localization~\citep{Lassiter1975ims}. 

\begin{figure}
\begin{centering}
\includegraphics[width=1\columnwidth]{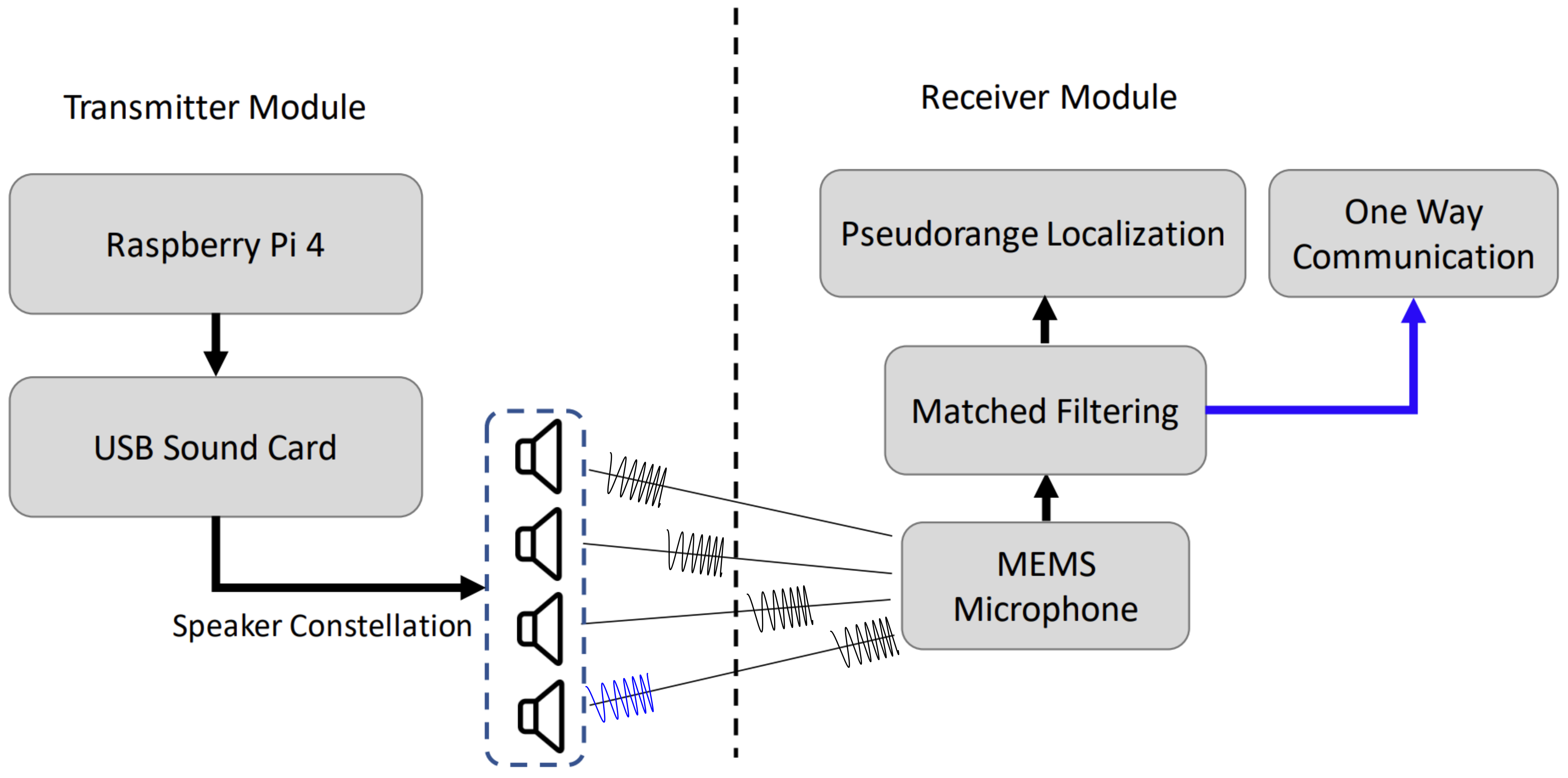}\caption{System Overview: The transmitter module periodically plays a sequence of identical chirps (black) followed by another, different, chirp (blue) through a constellation of speakers. The receiver module uses the information to estimate position as well as execute basic locomotion tasks.}\label{fig:system_overview}\vspace{-6mm}
\par\end{centering}
\label{fig:System-Overview-Loc}
\end{figure}

In this paper we present a method for localization of a small robotic fish that is based on acoustic pseudoranging.  This is accomplished using cheap, miniaturized, low power sensors and computation. Referring to Fig.~\ref{fig:system_overview}, pseudorange localization and acoustic communication is performed on a fish-inspired robot that swims in a water tank that is instrumented with acoustic speakers and receivers. The robotic fish is embedded with a small and inexpensive micro-electro-mechanical systems (MEMS) microphone that is used for communication and localization through the estimation of TDOA of signals sent simultaneously from the four speakers in the water tank. 

As an acoustically passive method on the receiver side, pseudoranging allows for multiple agents to be localized simultaneously without the need for individual signals specifying time of flight (TOF) information or cross communication between transmitter and receiver to achieve time of arrival (TOA). Variations of acoustic pseudoranging have been utilized for the localization of larger underwater vehicles before. Jorgensen et al.~\citep{Jorgensen2020joe} presented an observer to estimate several parameters like position, velocity and IMU biases. Leveraging pseudorange measurement differences along with attitude and accelerometer gave better position estimates. A long range underwater navigation algorithm based on acoustic pseudoranging was tested in an area spanning \textasciitilde 275,000$km^{2}$ by Mikhalevsky et al.~\citep{Mikhalevsky2020JASA} by using GPS assisted beacons. Recent works by Berlinger et al.~\citep{Berlinger2021} and Novak et al.~\citep{Novak2019} have also explored the localization of a swarm of fish, both robotic and natural, through optical and acoustic methods respectively. However both methods perform localization from an external observer and not on-board the agent, which makes acoustic pseudoranging more suitable to use cases where geo-referenced data collection, and on-board navigation are important.


\section{System Architecture\label{sec:System}}

This section will now describe the hardware and framework for acoustic pseudoranging based localization and one-way acoustic communication. A diagrammatic illustration is shown in Fig.~\ref{fig:system_overview}. There are a minimum of four speakers connected to a single computer playing a sequence of chirps followed by a single, different chirp, periodically. The receiver is time synchronized with the transmitter computer before deployment. The receiver, equipped with a MEMS microphone, records the acoustic signal from the transmitter's speakers, and processes the recorded sequence to obtain the pseudoranges to each speaker. The pseudorange observations are then used to solve for the receiver position and time bias between the receiver and transmitter clock. While an initial time sync is necessary to coordinate when the receiver should expect the signal, having a low-latency or highly accurate synchronization is not required. The same setup can also be used to listen to different types of signals to receive commands for certain tasks. Details on the system hardware and framework are discussed in detail in the subsequent subsections. 

\subsection{System Hardware}

\subsubsection{Acoustic Transmitters}

The transmitter module consists of four Lubell UW30 underwater speakers connected to a single Raspberry Pi 4 through a StarTech USB multi-channel sound card. The Raspberry Pi 4 is connected to a wireless network and makes available its local system time through network time protocol (NTP). A four channel WAV file with the same chirp signal in each channel, staggered at known intervals, is played periodically every 10 seconds. The chirp pattern used is a 10 ms, 4.5-8.5kHz linear up-chirp. While using four diverse chirps with different frequency bandwidths, played simultaneously would intuitively be more convenient, having a wider bandwidth and moderately longer chirp length gives better ranging resolution when using matched filters~\citep{Lazik2012Sensys}. Thus the linear up-chirp is repeated at intervals of 200 ms across the 4 channels, with the sequence order known beforehand.We also use three shorter chirps in different bandwidths to each other and the aforementioned sequence to communicate different motion behaviours for the robotic fish. One of these shorter chirps then follows the sequential chirps for the purpose of relaying motion instructions to the robot.

\subsubsection{Receiver and Bio-inspired Robot}

\begin{figure}[th]
\begin{centering}
\subfloat[Static receiver module]{\centering{}\includegraphics[width=0.95\columnwidth]{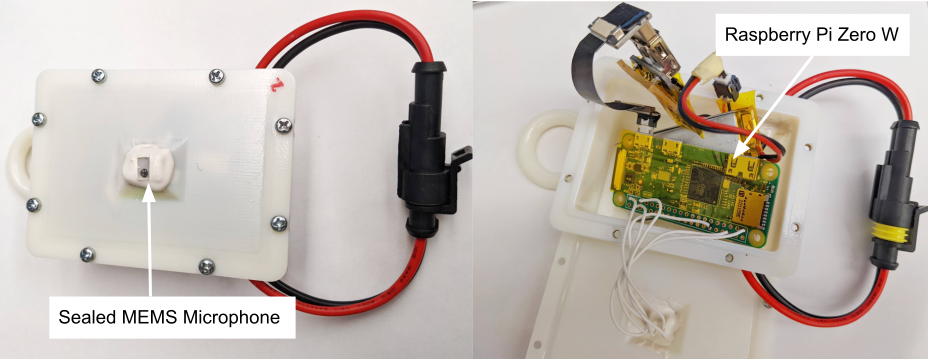}}\medskip{}
\subfloat[Receiver module on fish-inspired robot]{\centering{}\includegraphics[width=0.95\columnwidth]{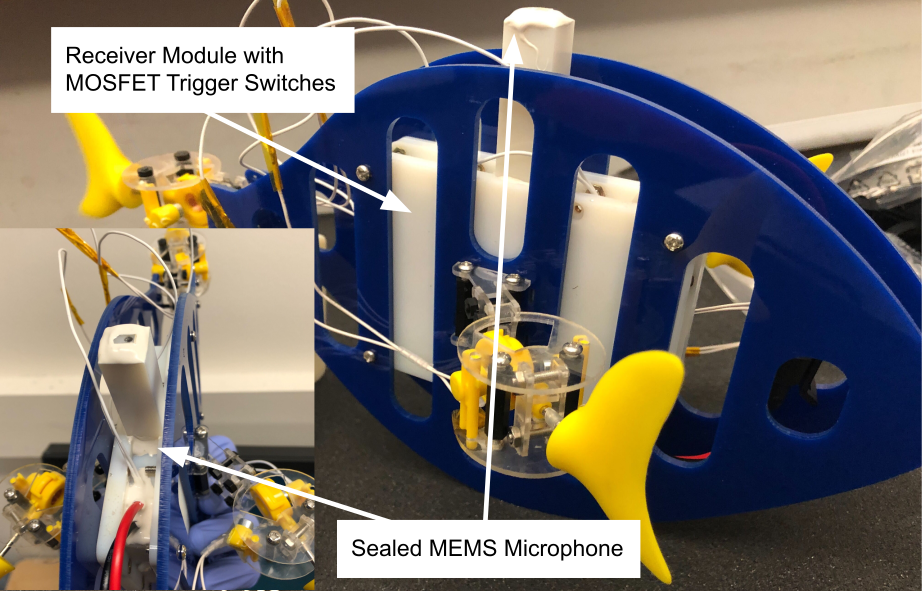}}
\par\end{centering}
\caption{Acoustic Receiver Setup: (a) Receiver module for localization tests: While the MEMS microphone is sealed for water tightness, its bottom port is only covered by a very thin membrane to reduce the dampening of incoming signals. (b) The proof-of-concept bio-inspired fish prepared for experiments is internally identical to the box used for localization evaluation with added MOSFET trigger switches to control the actuators for its fins. }\label{fig:Box}\vspace{-3mm}
\end{figure}

The receiver module runs on a Raspberry Pi Zero W powered by a 5V Li-Poly battery. The only sensor attached is the ICS43434 I2S MEMS microphone. The choice of using a MEMS microphone over commercially available hydrophones was made considering the significantly lower cost, and smaller form factor of the microphones. The receiver module is synchronized with the transmitter clock through NTP over wireless networks before submersion in water. After this point, the inbuilt RTC clock is sufficient to keep track of the periodic signal being sent by the transmitter. The receiver begins recording $2.5s$ long WAV files at each time step. The recording starts before the transmitter module plays the signals to keep ample buffer space to ensure all the chirps are recorded. Accumulated clock drift is corrected by using the time bias calculated in the pseudorange solution. The 1 GHz ARM11 32-bit processor of the Raspberry Pi Zero W takes approximately $6.05s$ for each final location estimate, including the recording duration. Fig.~\ref{fig:Box} (a) shows the static receiver module used for localization experiments. The bio-inspired fish robot used for experiments is shown in Fig.~\ref{fig:Box} (b). It functions on the same components as its static counterpart with the addition of MOSFET trigger switches for the fins. To achieve locomotion, each fin was individually activated by the commands from the controller to generate thrusts that result in overall forward or turning motions. All the electronic components and battery were waterproofed inside the central box while the cables carrying alternating current were connected to the end actuators at the fins.

\subsection{Localization}

The localization framework first performs the matched filtering of the recorded acoustic signal and uses the filter results for the pseudoranging based localization. As a pre-step, we also analyze speaker constellation geometry. The steps are explained in detail in the following sub-sections. 

\subsubsection{Data processing and matched filtering}

We perform matched filtering on the recorded chirp sequence from the four speakers to obtain our pseudorange estimates. The matched filter is a linear filter suited for maximizing the signal-to-noise ratio of the recorded signal. The filter conducts a convolution between the recorded sequence, $x[n]$ and a replica of the original chirp played by the speakers $h[n]$:
\begin{equation}
y[n]=\sum_{k=0}^{N-1}h[n-k]x[k]
\end{equation}

\begin{figure}
\begin{centering}
\includegraphics[viewport=0bp 0bp 1192bp 604bp,width=1\columnwidth]{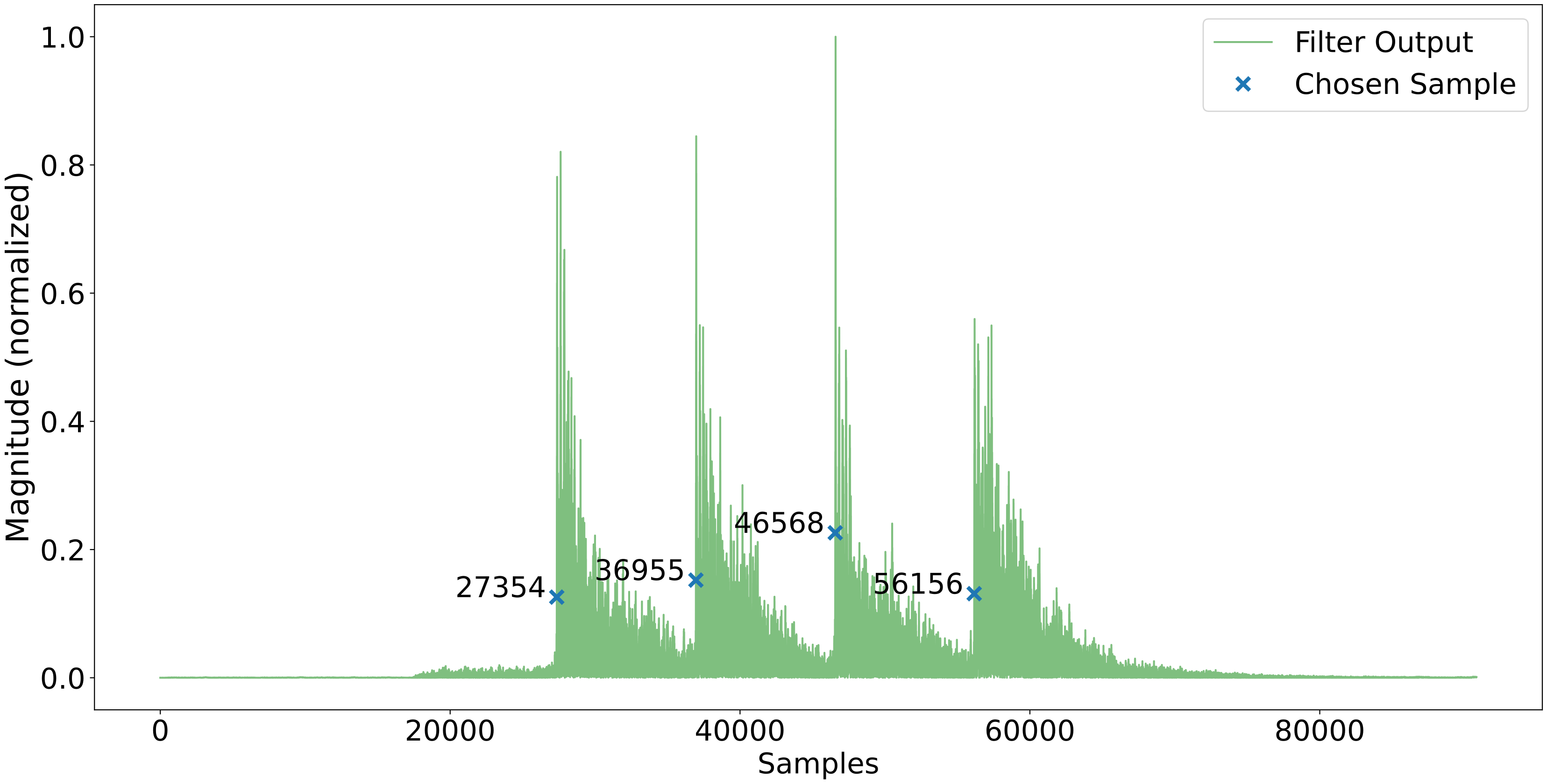}
\par\end{centering}
\caption{Matched Filter Example Output: The normalized filter output is compared against a threshold to pick the first significant peak for each chirp. The sample numbers, after adjusting for the known 9600 sample delay, provide a pseudorange of 843.415, 843.445, 843.847, and 843.477 meters to each speaker as in Eq.~\ref{eq:pr_obs}.}
\label{fig:MatchedFilter}
\end{figure}

The sample number that resembles the replica $h[n]$ most gives a maximal or near maximal amplitude for output $y[n${]}. Typically, the maximum output would indicate the sample number to be chosen. However, due to non-line-of-sight reflections in smaller enclosed environments, we have observed that choosing the first major peak gives the best results. The four pseudoranges are obtained by adjusting the known signal delay between each chirp, giving us the sample number for each chirp $s_{i}$ and multiplying by the speed of sound in fresh water (1480$\frac{m}{s}$) and dividing by the sampling rate (48kHz) gives us the pseudorange observation:
\begin{equation}
P^{i}=\frac{c}{F_{s}}s^{i}=\frac{1480}{48000}s^{i}\label{eq:pr_obs}
\end{equation} 
An example of the matched filter output is shown in Fig.~\ref{fig:MatchedFilter}. The filter output is normalized to keep a constant threshold to select the first significant peak for each chirp. As the expected gap between each chirp is known (9600 samples), the adjusted pseudorange observation for each speaker can be found and used to localize the receiver. 

\subsubsection{Acoustic Pseudoranging}

Acoustic pseudoranging is a TDOA technique inspired from GPS trilateration. Our implementation of acoustic pseudoranging differs from its GPS counterpart in a few ways. Unlike in the case of GPS satellites, the speaker positions are constant and known, as well as connected to the same central transmitting module. As such, it does not necessitate the need to account for transmitter side clock biases and errors. We refer to literature from GPS point positioning~\citep{Herring1999ieeeproc,Blewitt2000BasicsOT} estimation to formulate our acoustic pseudoranging equations which are described as follows:

Given speaker positions $[x^{i},y^{i},z^{i}]$ for speakers $i\in[1,4]$, receiver position $[x_{r},y_{r},z_{r}]$, time step $T$ and receiver clock bias $dt_{r}$, each pseudorange for the four or more speakers can be modeled as: 
\vspace{-2mm}
\begin{equation}
P^{i}(T)=R_{r}^{i}(T)+c\times dt_{r}(T)\label{eq:pseudorange}
\end{equation}
where the distance between each speaker $i$ and receiver $r$ is
given by the Euclidean distance
\begin{equation}
\resizebox{0.9\columnwidth}{!}{%
$R_{r}^{i}(T)=\sqrt{(x^{i}-x_{r}(T))^{2}+(y^{i}-y_{r}(T))^{2}+(z^{i}-z_{r}(T))^{2}}$\label{eq:Eucledian}
  }
\end{equation}

To find the location of the receiver with respect to the speakers, we need to solve for the unknown parameters which are the receiver position and receiver time bias $[x_{r},y_{r},z_{r},dt_{r}]$ at each time step $T$. Assuming initial estimates of receiver position and time bias, $[x_{0},y_{0},z_{0,}dt_{0}]$, are known, the relationship of the initial estimates of receiver $r$, and the update parameters can be written as: \vspace{-5mm}

\begin{align}
x_{r} & =x_{0}+\Delta x\nonumber, \hspace{4mm}
y_{r} =y_{0}+\Delta y\nonumber \\
z_{r} & =z_{0}+\Delta z, \hspace{3mm}
dt_{r}=dt_{0}+\Delta dt
\end{align}
$\Delta x,\Delta y,\Delta z,\Delta dt$ are now the parameters we use a least squares formulation to solve for. Using a first order Taylor expansion, Eq.~\ref{eq:pseudorange} can be expressed as
\begin{equation} \label{eq:taylor_exp}
\resizebox{0.9\columnwidth}{!}{%
        $
        P_{r}^{i}(T) =R_{0}^{i}(T)+\frac{\partial P}{\partial x}\Biggl|_{x_{0}}\negmedspace\negmedspace\negthickspace\Delta x+\frac{\partial P}{\partial y}\Biggl|_{y_{0}}\negmedspace\negmedspace\negthickspace\Delta y+\frac{\partial P}{\partial z}\Biggl|_{z_{0}}\negmedspace\negmedspace\negthickspace\Delta z+\frac{\partial P}{\partial dt}\Biggl|_{t_{0}}\negmedspace\negmedspace\negthickspace\Delta dt
        $
        }
\end{equation}

where $R_{0}^{i}$ is found from Eq.~\ref{eq:Eucledian} by using the initial position estimate. The partial derivatives from Eq.~\ref{eq:taylor_exp} are:

\begin{align}
\frac{\partial P}{\partial x}\Biggl|_{x_{0}} & \negmedspace\negmedspace\negthickspace=-\frac{x^{i}-x_{0}(T)}{R_{0}^{i}(T)}, \hspace{3mm} 
\frac{\partial P}{\partial y}\Biggl|_{y_{0}} \negthickspace=-\frac{y^{i}-y_{0}(T)}{R_{0}^{i}(T)}\nonumber \\
\frac{\partial P}{\partial z}\Biggl|_{z_{0}} & \negmedspace\negmedspace\negthickspace=-\frac{z^{i}-z_{0}(T)}{R_{0}^{i}(T)}, \hspace{3mm} 
\frac{\partial P}{\partial dt}\Biggl|_{t_{0}}\negthickspace=c\label{eq:partialD}
\end{align}

Using our partial derivatives from Eq.~\ref{eq:partialD}, and rearranging the known terms together, we can rewrite Eq.~\ref{eq:taylor_exp} as:
\begin{multline}
P_{r}^{i}(T)-R_{0}^{i}(T)=-\frac{x^{i}-x_{0}(T)}{R_{0}^{i}(T)}\Delta x-\frac{y^{i}-y_{0}(T)}{R_{0}^{i}(T)}\Delta y\\
-\frac{z^{i}-z_{0}(T)}{R_{0}^{i}(T)}\Delta z+c\times\Delta dt\label{eq:pseudorange-linear}
\end{multline}
For ease of reference, let:
\begin{align}
a_{x_{r}}^{i} & =-\frac{x^{i}-x_{o}(T)}{R_{o}^{i}(T)}\nonumber \\
a_{y_{r}}^{i} & =-\frac{y^{i}-y_{o}(T)}{R_{o}^{i}(T)}\nonumber \\
a_{z_{r}}^{i} & =-\frac{z^{i}-z_{o}(T)}{R_{o}^{i}(T)}\nonumber \\
b^{i} & =P_{r}^{i}(T)-R_{o}^{i}(T)
\end{align}
Assuming we have the minimum of four required speakers, we have the
system of linearized simultaneous equations:
\begin{align}
b^{1} & =a_{x_{r}}^{1}\Delta x+a_{y_{r}}^{1}\Delta y+a_{z_{r}}^{1}\Delta z+c\Delta dt\nonumber \\
b^{2} & =a_{x_{r}}^{2}\Delta x+a_{y_{r}}^{2}\Delta y+a_{z_{r}}^{2}\Delta z+c\Delta dt\nonumber \\
b^{3} & =a_{x_{r}}^{3}\Delta x+a_{y_{r}}^{3}\Delta y+a_{z_{r}}^{3}\Delta z+c\Delta dt\nonumber \\
b^{4} & =a_{x_{r}}^{4}\Delta x+a_{y_{r}}^{4}\Delta y+a_{z_{r}}^{4}\Delta z+c\Delta dt\label{eq:expanded}
\end{align}
which can be simplified in their matrix form as:
\begin{align}
A & =\left[\begin{array}{cccc}
a_{x_{r}}^{i} & a_{x_{r}}^{i} & a_{x_{r}}^{i} & c\\
a_{x_{r}}^{i} & a_{x_{r}}^{i} & a_{x_{r}}^{i} & c\\
a_{x_{r}}^{i} & a_{x_{r}}^{i} & a_{x_{r}}^{i} & c\\
a_{x_{r}}^{i} & a_{x_{r}}^{i} & a_{x_{r}}^{i} & c
\end{array}\right]\nonumber \\
\boldsymbol{x} & =\left[\begin{array}{c}
\Delta x\\
\Delta y\\
\Delta z\\
\Delta dt
\end{array}\right],\boldsymbol{b}=\left[\begin{array}{c}
b^{1}\\
b^{2}\\
b^{3}\\
b^{4}
\end{array}\right]\label{eq:matrix_form}
\end{align}
where:

$\boldsymbol{b}=$ prediction error

$\boldsymbol{x}=$ vector of unknowns which are the receiver position and clock bias update parameters.

$A=$ design matrix constructed from the linear functions of the unknown variables. 

\noindent Eq.~\ref{eq:expanded} can now be expressed in matrix-vector form as: 
\begin{equation}
A\boldsymbol{x}=\boldsymbol{b}
\end{equation}

To better represent the observed measurement errors, which in acoustic pseudoranging arises from selecting the right peak from our matched filter result, we would need to adjust the prediction error and design matrices. This can be done through a form of whitening as explained in~\citep{Dellaert17fnt}. Let $\Sigma=diag(\sigma_{1}^{2},\sigma_{2}^{2},\sigma_{3}^{2},\sigma_{4}^{2}$) where $\sigma_{1},..,\sigma_{4}$ represent the average error observed in pseudorange measurement for each speaker. The design and prediction error matrix are weighted as: 
\begin{equation}
A_{w}=\Sigma^{-1/2}A,B_{w}=\Sigma^{-1/2}\boldsymbol{b}
\end{equation}
Finally, our least squares problem is solved as:
\begin{equation}
\boldsymbol{x}=(A_{w}^{T}A_{w})^{-1}A_{w}^{T}B_{w}
\end{equation}
The updated parameters from $\boldsymbol{x}$ are then used to re-initialize the estimated receiver position. The least squares optimization is repeated until the desired error threshold is reached. 

\subsubsection{Dilution of Precision}

Akin to GPS, the ability of acoustic pseudoranging to provide an accurate position estimate depends on the geometry composed by the speaker positions. One of the most popular metrics is the geometric dilution of precision (GDOP). It can be obtained from the design matrix $A$ in Eq. \ref{eq:matrix_form} by getting the covariance matrix as:
\begin{equation}
\Sigma_{\boldsymbol{x}}=(A^{T}A)^{-1}\label{eq:covariance}
\end{equation}
The GDOP is given as the root of the trace of the covariance matrix,
$\Sigma_{\boldsymbol{x}}$~\citep{Yarlagadda2000Radar}:
\begin{equation}
GDOP=\sqrt{tr(\Sigma_{x})}\label{eq:GDOP}
\end{equation}
A lower GDOP value indicates a better constellation geometry. Based on an analysis by Isik et al.~\citep{Isik2020robotics}, values between 0--5 are considered to give excellent to great estimates, 5--10 give moderate and values 10--20 give a fair estimate. Values above are unlikely to give a reliable solution. As such, the use of acoustic pseudoranging in small spaces requires speaker geometry configurations be analyzed beforehand to ensure good estimation accuracy is obtained. 

\subsection{Communication}

Apart from acoustic localization, the same hardware and framework can be leveraged to perform one way communication to perform some pre-determined behaviours. As seen earlier in Section \ref{sec:Introduction-and-Related} the controller used by Katzschmann et al.~\citep{Katzschmann18science,DelPreto2015wuwnet} demonstrated the acoustic control of bio-inspired robots while in water. While this method is highly suitable for precise control of a single robot, its ability to simultaneously pass commands to several robots is to be seen. Fischell et al.~\citep{FischellRAL2019} present a more viable solution for multiple robots to be given acoustic commands through a single speaker. While, the current framework with a single microphone will be unable to perform maneuvers needing bearing estimates to the beacon, it is still possible to give general commands relating to diving or re-surfacing, and directional commands when equipped with other supplementary sensors like IMUs or depth sensors. In our work, as a proof-of-concept, we execute directional commands to move a robotic fish with three actuated fins. We once again rely on matched filtering to differentiate between the type of communication signal received as well as the sequence used for localization. The maximal filter output determines the motion to be executed. The chirps used for communication can have shorter frequency bandwidths since timing resolution is not important to perform simple detection. 

\section{Evaluation}

To test the localization framework, experiments were performed in a cylindrical water tank with diameter 7.3m and depth 2.7m. The tests were performed using a water-tight box containing the receiver module described in Section \ref{sec:System} as seen in Fig \ref{fig:Box}. Ground truth positions were measured by using a graduated rope across the tank with the receiver hanging down from marked positions at pre-measured heights from the bottom of the tank as in Fig.~\ref{fig:experiment_pics} (a). 

\begin{figure*}[t]
\begin{centering}
\subfloat[Configuration 1]{\centering{}\includegraphics[clip,width=0.32\textwidth]{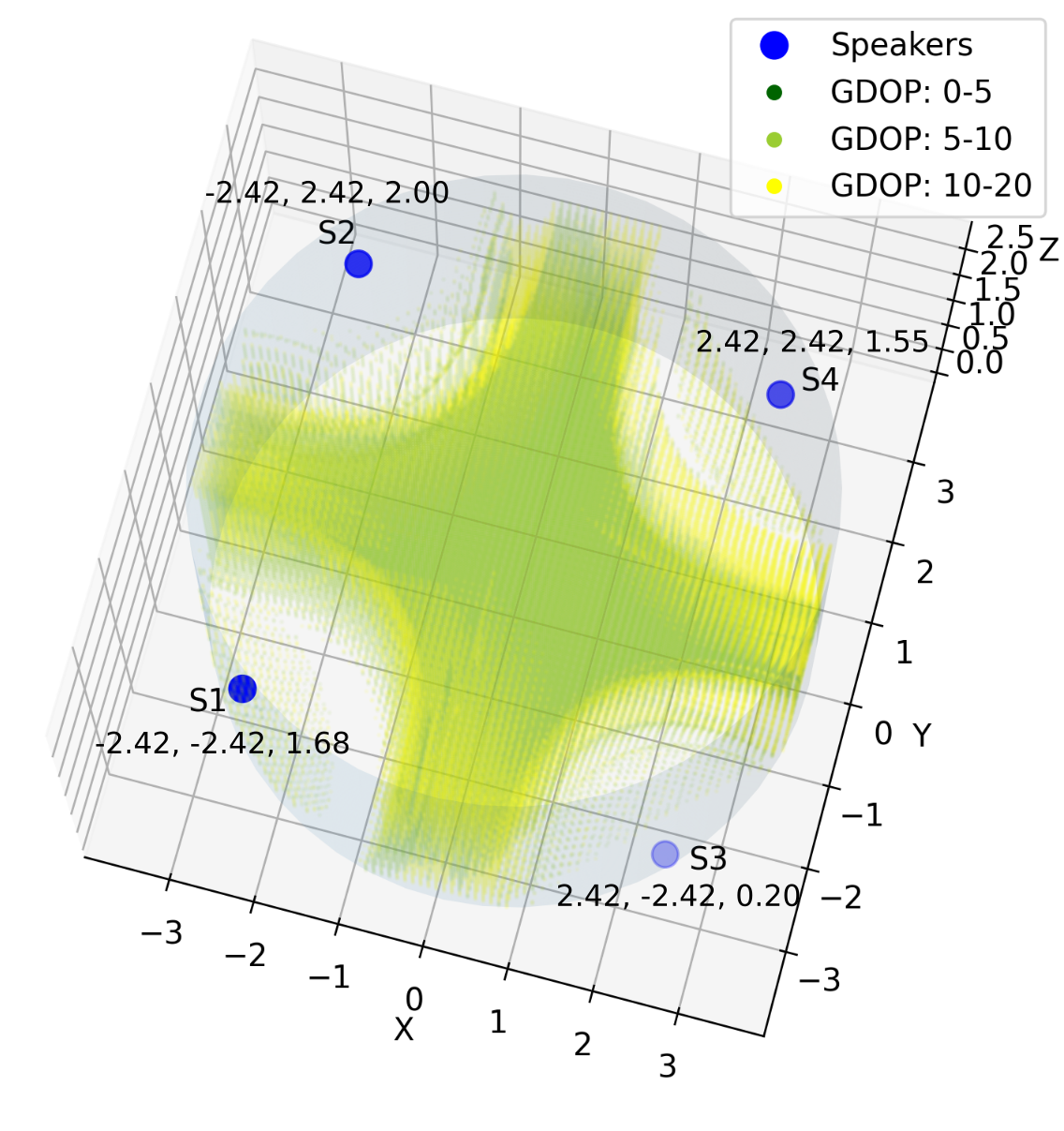}}\subfloat[Configuration 2]{\centering{}\includegraphics[width=0.32\textwidth]{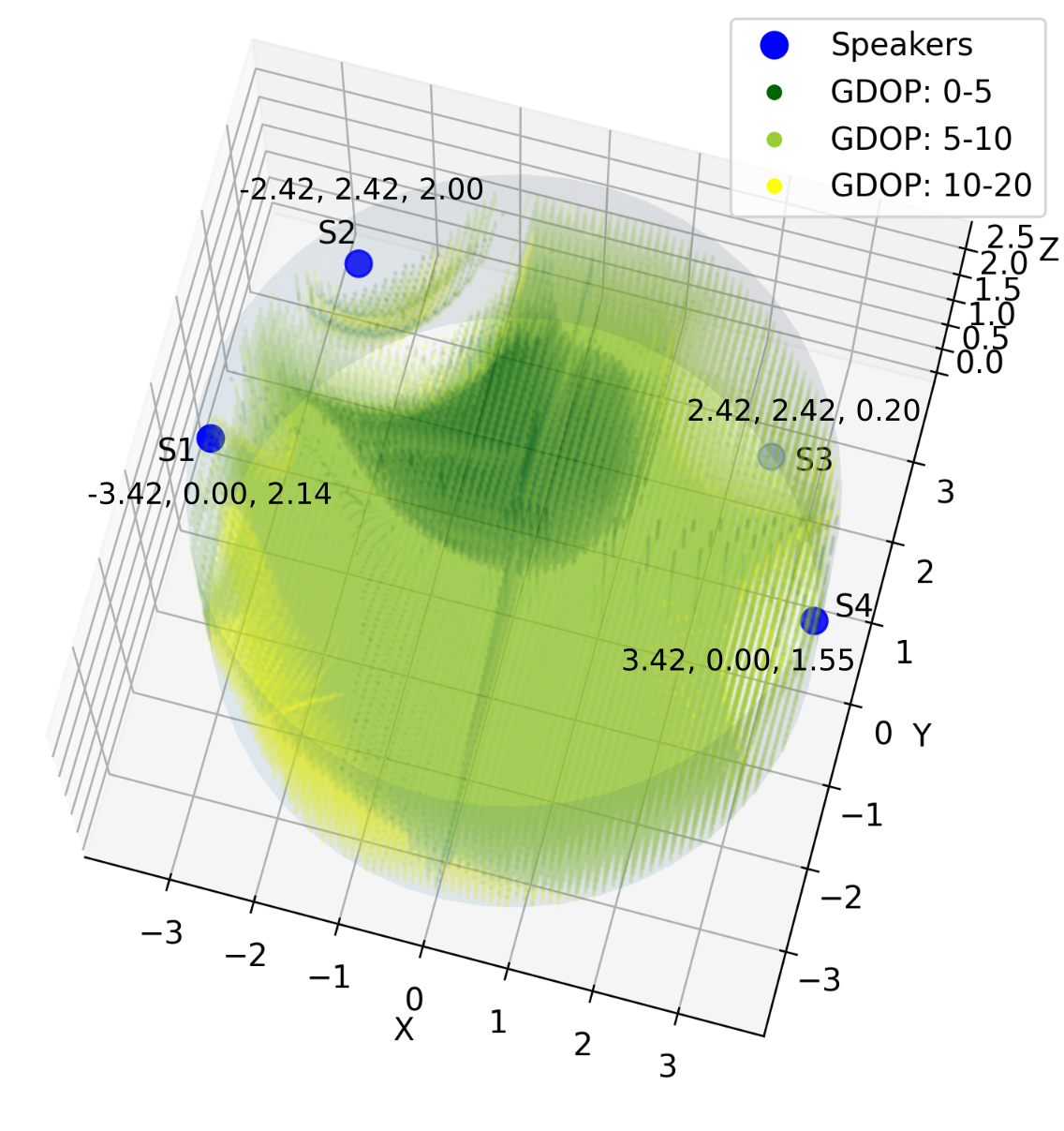}}\subfloat[Configuration 3]{\centering{}\includegraphics[clip,width=0.32\textwidth]{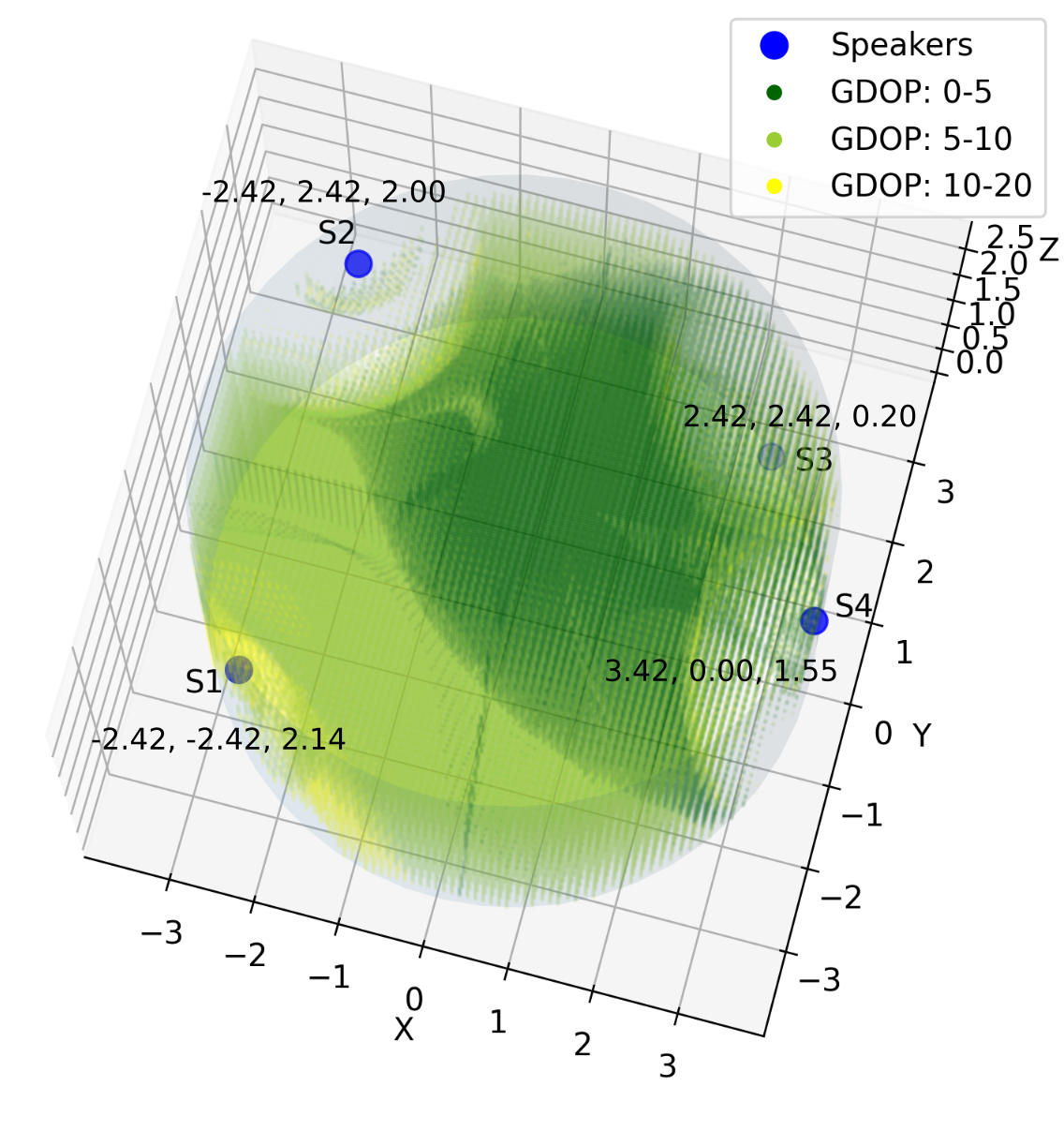}}
\par\end{centering}
\caption{Analysis Of Available Volume: The three configurations show the solvable volume ranging from yellow to dark green, depicting a decrease in GDOP value. All points were provided an initial position estimate of (0,0,1). Configuration 1 shows the lowest solvable volume despite a better spread of the speakers in the X-Y plane. However due to the smaller variation in the speaker heights, a good spread in the X-Y plane is not enough to give reliable estimates. The average GDOP value for the solvable space for each configuration was calculated to be 9.39, 6.81 and 5.54 respectively, as evidenced with the increase in the dark green points as seen in (b) and (c). }
\label{fig:gdop_analysis}
\end{figure*}

The experiments were repeated in three different speaker geometry configurations for a comparative study. First, the total available volume for each configuration was analysed to obtain their average GDOP values and "solvable volume". We divide our analysed space into four groups, with GDOP values ranging from 0--5, 5--10, 10--20 and lastly, unsolvable space, where we are unlikely to receive a good estimate of the receiver position. Configuration 1 is indicative of a moderate speaker geometry where only 59\% of the total available volume of the tank is able to give a GDOP value under 20. Configurations 2 and 3 on the other hand prove to be better constellation geometries with 88\% of the total available volume having GDOP value under 20. Details on speaker positions and further information are presented
in Fig.~\ref{fig:gdop_analysis}.

\begin{figure*}[t]
\begin{centering}\vspace{-5mm}
\subfloat[Configuration 1]{\centering{}\includegraphics[width=0.205\paperwidth]{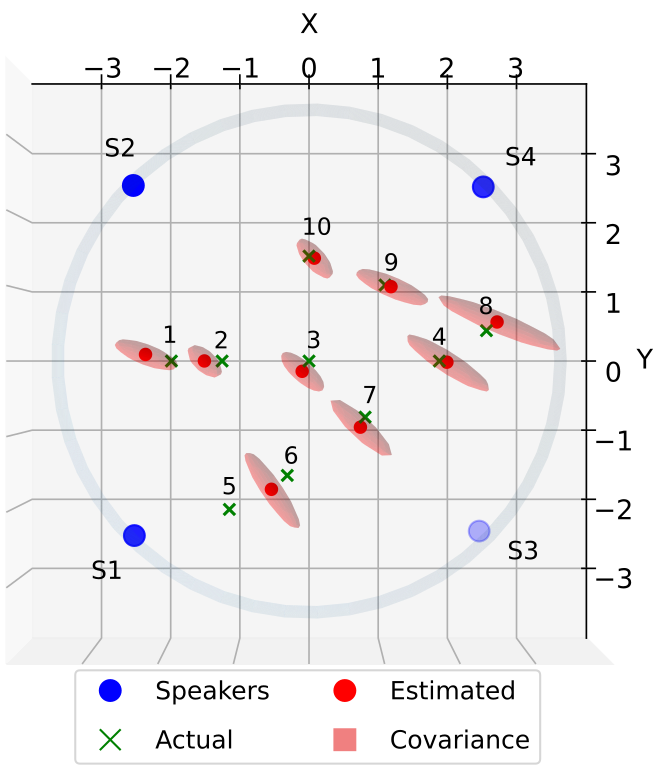}}\subfloat[Configuration 1: Depth Variation]{\centering{}\includegraphics[width=0.205\paperwidth]{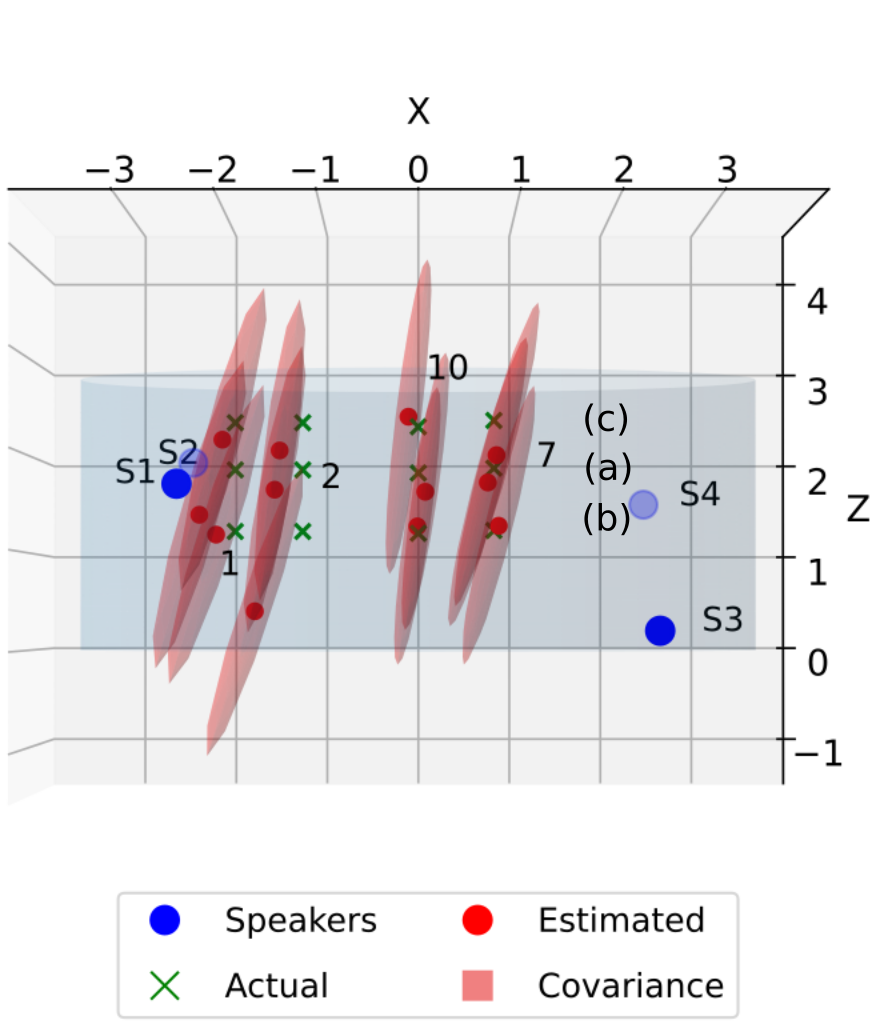}}\subfloat[Configuration 2]{\centering{}\includegraphics[width=0.205\paperwidth]{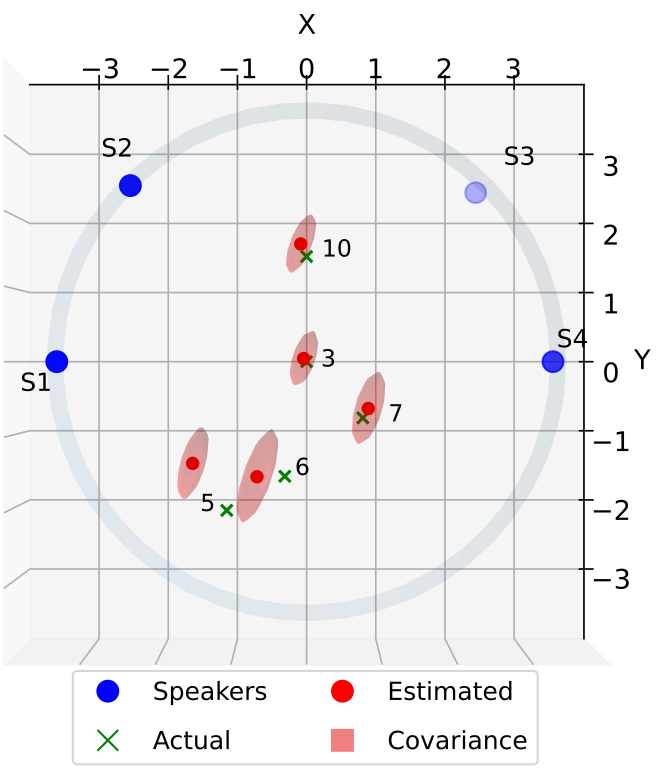}}\subfloat[Configuration 3\label{fig:Configuration-3}]{\centering{}\includegraphics[width=0.205\paperwidth]{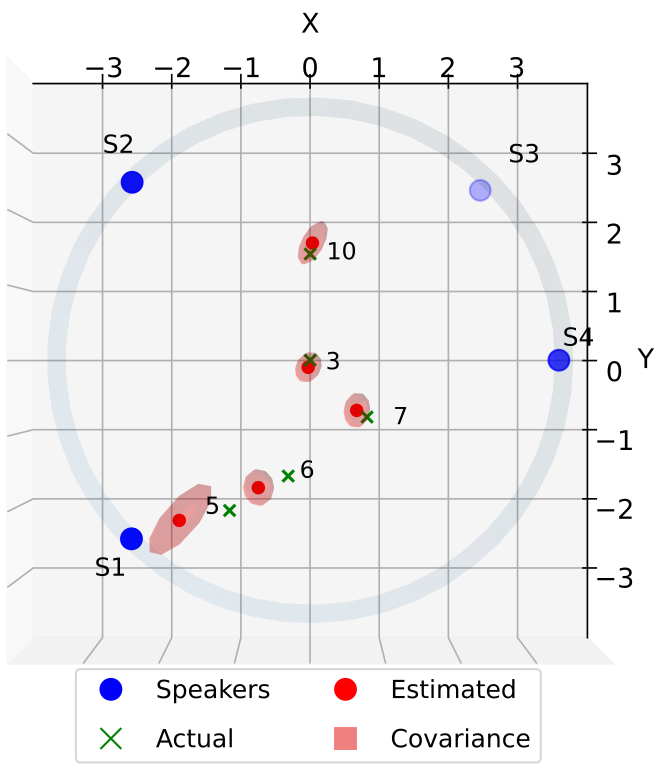}}
\par\end{centering}
\caption{Localization Results: Localization estimates for the three configurations are presented. All positions were provided with a starting estimate of (0,0,1). (a) Configuration 1 achieves good estimates despite the weaker constellation geometry. However, being in an unsolvable space, like position 5, would give a very poor estimate. (b) The largest variation in speaker depth is between speaker 2 and 3 at \textasciitilde 2m. This causes narrower elevation angles to the receiver making it difficult to achieve lower variance in depth estimates. (c) and (d) compare the change in position accuracy and confidence for the same positions with only speaker 1 shifting position. We see that having more variation is beneficial to achieve better localization accuracy.}
\label{fig:Loc_results}
\end{figure*}\vspace{-4mm}

\begin{table}[th]
\centering{}\caption{Configuration 1 Results}
\begin{tabular}{|c|c|c|c|c|}
\hline 
Pos No. & Ground Truth & Estimate & MSE & XY RMSE\tabularnewline
\hline 
\hline 
1 & -1.90,~0.00,~1.87 & -2.27,~0.09,~1.40 & 0.605 & 0.387\tabularnewline
\hline 
2 & -1.20,~0.00,~1.87 & -1.44,~0.00,~2.07 & 0.313 & 0.239\tabularnewline
\hline 
3 & ~0.00,~0.00,~1.87 & -0.09,-0.14,~1.73 & 0.225 & 0.171\tabularnewline
\hline 
4 & ~1.80,~0.00,~1.87 & ~1.91,-0.02,~1.61 & 0.288 & 0.114\tabularnewline
\hline 
5 & -1.10,-2.05,~1.87 & Out of Bounds & NA & NA\tabularnewline
\hline 
6 & -0.30,-1.58,~1.87 & -0.52,-1.77,~1.86 & 0.292 & 0.292\tabularnewline
\hline 
7 & ~0.77,-0.77,~1.87 & ~0.71,-0.92,~1.72 & 0.217 & 0.157\tabularnewline
\hline 
8 & ~2.45,~0.42,~1.87 & ~2.60,~0.54,~1.74 & 0.234 & 0.195\tabularnewline
\hline 
9 & ~1.05,~1.05,~1.87 & ~1.13,~1.03,~1.92 & 0.100 & 0.084\tabularnewline
\hline 
10 & ~0.00,~1.45,~1.87 & ~0.07,~1.43,~1.67 & 0.214 & 0.077\tabularnewline
\hline 
  \textbf{Mean} &&& \textbf{0.276} & \textbf{0.190}\tabularnewline
\hline 
\end{tabular}\label{RMSE_table-1-1}\vspace{0cm}
\end{table}


\begin{table}[th]
\centering{}\caption{Configuration 1 Depth Variation Results}
\begin{tabular}{|c|c|c|c|c|}
\hline 
Pos No. & Ground Truth & Estimate & RMSE &XY RMSE\tabularnewline
\hline 
\hline 
1 a & -1.90,~0.00,~1.87 & -2.27,~0.09,~1.40 & 0.605 & 0.387\tabularnewline
\hline 
1 b & -1.90,~0.00,~1.23 & -2.10,~0.04,~1.19 & 0.207 & 0.205\tabularnewline
\hline 
1 c & -1.90,~0.00,~2.36 & -2.03,~0.03,~2.18 & 0.224 & 0.139\tabularnewline
\hline 
2 a & -1.20,~0.00,~1.87 & -1.44,~0.00,~2.07 & 0.313 & 0.239\tabularnewline
\hline 
2 b & -1.20,~0.00,~1.23 & -1.70,~0.04,~0.40 & 0.967 & 0.499\tabularnewline
\hline 
2 c & -1.20,~0.00,~2.36 & -1.49,~0.11,~1.67 & 0.761 & 0.314\tabularnewline
\hline 
7 a & ~0.77,-0.77,~1.87 & ~0.71,-0.92,~1.72 & 0.217 & 0.157\tabularnewline
\hline 
7 b & ~0.77,-0.77,~1.23 & ~0.82,-0.83,~1.28 & 0.09 & 0.077\tabularnewline
\hline 
7 c & ~0.77,-0.77,~2.36 & ~0.80,-0.79,~2.00 & 0.358 & 0.032\tabularnewline
\hline 
10 a & ~0.00,~1.45,~1.87 & ~0.07,~1.42,~1.67 & 0.214 & 0.077\tabularnewline
\hline 
10 b & ~0.00,~1.45,~1.23 & -0.01,~1.60,~1.30 & 0.170 & 0.153\tabularnewline
\hline 
10 c & ~0.00,~1.45,~2.36 & -0.10,~1.59,~2.47 & 0.205 & 0.172\tabularnewline
\hline 
\textbf{Mean} &&& \textbf{0.360} & \textbf{0.204}\tabularnewline
\hline 
\end{tabular}\label{RMSE_table-1-2}\vspace{0cm}
\end{table}



A diverse number of positions were chosen to obtain localization estimates. Mean error across all trials was found to be \textbf{0.345m} with a standard deviation of \textbf{0.228m}. We break down our results on the basis of the speaker geometry configuration. Configuration 1 had tests conducted for several X-Y coordinates in the tank as well as static X-Y position with different depth placements. Fig.~\ref{fig:Loc_results} shows the estimated positions based on pseudorange observations along with covariance ellipses drawn to the confidence level of 2$\sigma$. As seen in Fig.~\ref{fig:Loc_results} (b) the covariance in Z or depth exceeds the covariance in X-Y. This is also seen in Tables \ref{RMSE_table-1-1}, \ref{RMSE_table-1-2}, \ref{RMSE_table-c2} and \ref{RMSE_table-c3}, where root mean square error (RMSE) in the X-Y plane is considerably lower than the total RMSE. This is explained by the smaller variation (\textasciitilde 2m) of the depths in the speaker constellation geometry compared to the variation in the X-Y positions (\textasciitilde 7m).

The localization results shown in Fig~\ref{fig:Loc_results} (a) and (b) for configuration 1 are tabulated in Table \ref{RMSE_table-1-1} and Table \ref{RMSE_table-1-2}. To be noted, position 5 is in an area which is considered to be unsolvable as per our GDOP analysis from Fig.~\ref{fig:gdop_analysis} (a). As predicted by the analysis, experimental results for position 5 provide an out of bounds estimate. Comparatively, configurations 2 and 3 have a larger solvable volume, and are able to give reasonable localization estimate for position 5, although with relatively higher error than other points. Tables \ref{RMSE_table-c2} and \ref{RMSE_table-c3} for configurations 2 and 3 show results for tests conducted at the same positions which can be visualized in Fig.~\ref{fig:Loc_results} (c) and (d). Comparing Fig.~\ref{fig:Loc_results} (c) and (d) shows that larger variance in speaker positions is likely to give smaller localization error as evidenced by the smaller covariance ellipses seen for configuration 3. We ignore errors through multi-path reflections and speaker washout (being close enough to any speaker may produce artifacts in recorded signal sequences). We still find the localization results to be satisfactory considering the low cost of implementing such a system.

Acoustic communication is successfully performed to cycle through three different motions using matched filtering to differentiate between the acoustic signals. The motions we implemented include moving the tail fin alone to move forward, and move left or right by simultaneously moving the tail and appropriate pectoral fin.

\begin{table}[th]
\centering{}\caption{Configuration 2 Results}
\begin{tabular}{|c|c|c|c|c|}
\hline 
Pos No. & Ground Truth & Estimate & RMSE & XY RMSE\tabularnewline
\hline 
\hline 
3 & ~0.00,~0.00,~1.87 & -0.04,~0.04,~1.98 & 0.128 & 0.059\tabularnewline
\hline 
5 & -1.10,-2.05,~1.87 & -1.56,-1.39,~2.18 & 0.745 & 0.677\tabularnewline
\hline 
6 & -0.30,-1.58,~1.87 & -0.69,-1.60,~1.46 & 0.569 & 0.391\tabularnewline
\hline 
7 & ~0.77,-0.77,~1.87 & ~0.84,-0.64,~2.14 & 0.312 & 0.149\tabularnewline
\hline 
10 & ~0.00,~1.45,~1.87 & -0.08,~1.62,~1.79 & 0.211 & 0.194\tabularnewline
\hline 
 \textbf{Mean} &&& \textbf{0.393} & \textbf{0.294}\tabularnewline
\hline 
\end{tabular}\label{RMSE_table-c2}\vspace{0cm}
\end{table}


\begin{table}[th]
\centering{}\caption{Configuration 3 Results}
\begin{tabular}{|c|c|c|c|c|}
\hline 
Pos No. & Ground Truth & Estimate & RMSE & XY RMSE\tabularnewline
\hline 
\hline 
3 & ~0.00,~0.00,~1.87 & -0.03,-0.10,~1.71 & 0.186 & 0.101\tabularnewline
\hline 
5 & -1.10,-2.05,~1.87 & -1.81,-2.22,~1.26 & 0.754 & 0.448\tabularnewline
\hline 
6 & -0.30,-1.58,~1.87 & -0.71,-1.75,~1.65 & 0.495 & 0.445\tabularnewline
\hline 
7 & ~0.77,-0.77,~1.87 & ~0.63,-0.69,~1.81 & 0.168 & 0.159\tabularnewline
\hline 
10 & ~0.00,~1.45,~1.87 & ~0.03,~1.59,~1.86 & 0.153 & 0.153\tabularnewline
\hline 
 \textbf{Mean} &&& \textbf{0.351} & \textbf{0.261}\tabularnewline
\hline 
\end{tabular}\label{RMSE_table-c3}\vspace{0cm}
\end{table}


\section{Conclusion and Future Work}

This paper presents an acoustic localization and communication method suitable for small underwater robots such as bio-inspired robotic fish. Our approach uses a cheap MEMS microphone on the robot in combination with a constellation of four speakers in a known geometry that makes it possible for multiple robots to localize themselves simultaneously, which is useful for geo-referencing collected data. We present an implementation using a cost and power efficient Raspberry Pi Zero W on a bio-inspired robotic fish.

Enclosed spaces, like the water tank are prone to artifacts like multi-path reflections, speaker washout, and difficulty in finding ideal speaker constellation geometry. As such, acoustic pseudoranging is likely to perform better in larger, open waters. We show that despite being deployed in a small area, it is still effective in providing a reasonable position estimate. Future directions for the work on acoustic localization would involve the implementation of filtering and smoothing for trajectory estimation and optimization. Information from auxiliary sensors like IMUs and MEMS pressure sensors when fused with pseudoranging-based position estimates can aid the robot to have a more accurate state estimate. We aim to prove the efficacy of the improved method in open water compared to traditional inertial navigational methods used for underwater robots. 

The paper also demonstrates the use of the same architecture for controlling a robot through acoustic signals. While we limit our experiments to basic signal processing and primitive motions, more robust communication protocols like frequency shift keying can be used to perform a vast array of motions and tasks. Future iterations would aim to have a more sophisticated communication framework to achieve closed loop navigation and control, pushing bio-inspired underwater robots further towards autonomy. 

\vspace{1.8cm}
{\footnotesize\bibliographystyle{ieeetr}
\bibliography{references}}

\end{document}